\documentclass[11pt,a4paper]{article}
\usepackage[hyperref]{acl2018}
\usepackage{times}

\usepackage{latexsym}
\usepackage{url}

\usepackage{float}
\usepackage{amsmath}
\usepackage{amssymb}
\usepackage{color}
\usepackage{booktabs}
\usepackage{multirow}
\usepackage{enumitem}
\usepackage{graphicx}
\usepackage[most]{tcolorbox}
\usepackage{rotating}
\usepackage{placeins}
\usepackage{nomencl}
\usepackage{subcaption}

\newsavebox\mysavebox

\newtcolorbox[auto counter,number within=chapter]{samplebox}[1][]{
	enhanced,
	sharpish corners,
	fonttitle=\bfseries, 
	colframe=gray,
	title={Sample \thetcbcounter},
	#1
}

\aclfinalcopy 

\title{Robust and Scalable Differentiable Neural Computer\\for Question Answering}

\author{
  J\"org Franke, Jan Niehues, Alex Waibel \\
  Institute for Anthropomatics and Robotics \\
  Karlsruhe Institute of Technology, Germany \\
  {\tt joerg.franke@student.kit.edu} \\ 
  {\tt \{jan.niehues,alex.waibel\}@kit.edu} \\}

\date{}

\begin{document}
\maketitle

\begin{abstract}

Deep learning models are often not easily adaptable to new tasks and require task-specific adjustments. The differentiable neural computer (DNC), a memory-augmented neural network, is designed as a general problem solver which can be used in a wide range of tasks. But in reality, it is hard to apply this model to new tasks. 
We analyze the DNC and identify possible improvements within the application of question answering. This motivates a more robust and scalable DNC (rsDNC). The objective precondition is to keep the general character of this model intact while making its application more reliable and speeding up its required training time. 
The rsDNC is distinguished by a more robust training, a slim memory unit and a bidirectional architecture.
We not only achieve new state-of-the-art performance on the bAbI task, but also minimize the performance variance between different initializations. 
Furthermore, we demonstrate the simplified applicability of the rsDNC to new tasks with passable results on the CNN RC task without adaptions.

\end{abstract}

\section{Introduction}

In contrast to traditional statistical models, which often require a large amount of human effort on feature engineering and task-specific adjustments, a promise of deep learning is that little task-specific knowledge and minimal adaption is required to achieve state-of-the-art performance on different tasks.
But in reality, many deep learning solutions have to be adapted to a specific task to achieve good performance.  

However, there are more universal approaches for example the differentiable neural computer (DNC). 
It is introduced by \newcite{graves2016hybrid} as a general artificial neural network (ANN) model with an external memory ``to solve complex, structured tasks". 
It can be seen as a generic memory-augmentation framework. Unlike a vanilla ANN, it separates computation and memorization with a computational controller and a memory unit, which are independently modifiable. 
This allows a more accurate model design. Due to its fully differentiable design, it can be learned in a supervised fashion. 

The original paper shows applications on the bAbI question answering (QA) task, graph experiments and a reinforcement learning block puzzle solver \cite{graves2016hybrid}. But when applying this model to new QA tasks, no satisfying results are achieved. The issues of QA are the huge vocabulary, the length of the contexts and the required model complexity to find the correct answer. 

In this work, we analyze the DNC in QA tasks and identify four main challenges:
1. High memory consumption makes it hard to train large models efficiently. 2. The large variance in training performance within different initializations. 3. A slow and unstable convergence requires long varying training times. 4. The unidirectional architecture makes it hard to handle variable question appearance.

This work addresses these issues while keeping the general character of the model intact. We extend the DNC to be more robust and scalable (rsDNC) with the following contributions: 

\begin{enumerate}[topsep=0pt,itemsep=-1ex,partopsep=1ex,parsep=1ex]
\item A robust training with a strong focus on an early memory usage and normalization.
\item The usage of a slim, memory efficient, content-based memory unit for QA tasks.
\item A bidirectional DNC which allows a richer encoding of the input sequences. 
\end{enumerate}

The rsDNC is evaluated on two datasets. 
On the synthetic bAbI task \cite{weston2015towards}, we show performance improvements by 80\% compared to the DNC. These are new state-of-the-art results within multiple runs in a joint training. 
We also decrease the variance by up to 90\% between different random initializations.
Additionally, with training-data augmentation on one task, our model solves all tasks and provides the best-recorded results to the best of our knowledge. 
On the CNN RC task \cite{hermann2015teaching}, we show the adaptability of the rsDNC and achieve passable results without task-specific adaption.

\section{Related Work}

This section considers the related work regarding the two used datasets.

\paragraph{Related to bAbI task}

\newcite{rae2016scaling} provides technical enhancements with the introduction of the sparse DNC (SDNC) with sparse read and write operations. They allow to modifying only a sparse subset of interesting memory locations instead of manipulating all. This renders the memory consumption independent of memory size. 
A different approach provides the dynamic memory network (DMN) and its successor the DMN+ \cite{kumar2016ask,xiong2016dynamic}. They store sentence representations in an episodic memory and use attention to find the correct answer. 

The relation memory network (RMN) embeds sentences into a memory object and applies multiple times attention to find the answer \cite{yang2018finding}. In contrast to our model, the DMN+ and the RMN are optimized for QA tasks, uses sentence representation and require a dedicated question. The recurrent entity network (EntNet) ``can be viewed as a set of separate recurrent models whose hidden states store the memory slots" \cite{henaff2016tracking}. The memory slots or locations consist of a key vector and a content vector and have their own gated RNN as a controller. In contrast, our model has one memory matrix with no distinction between key and content.

\paragraph{Related to CNN RC task}
\newcite{hermann2015teaching} introduce the Deep LSTM Reader and the Attentive Reader which build a document representation by direct attention. 
\newcite{chen2016thorough} introduce the Stanford Attentive Reader which enhances the attentive reader and adds a bilinear term to compute the attention between document and query. The Attention-Sum (AS) Reader from \newcite{kadlec2016text} also uses separate encoding for the document and the query. Its successor, the Attention-over-Attention (AoA) Reader, applies a two-way attention mechanism to find the answer \cite{cui2017attention}. 

The ReasoNet uses iterative reasoning over a hidden representation of the document \cite{shen2017reasonet}.  
The Gated-Attention (GA) Reader from \newcite{dhingra2017gated} uses multiple hops over the document to build an attention over the candidates to select the answer token. 
These models are all conceptually adapted to the QA tasks they solve. In contrast, our solution is more versatile due to a more flexible and universal design. The different parts of the DNC can be exchanged or adjusted independently which allows simpler handling of new tasks. The versatility is shown in the original paper with three different tasks \cite{graves2016hybrid}.


\begin{figure}[!t]
	\centering
	\includegraphics[scale=0.1]{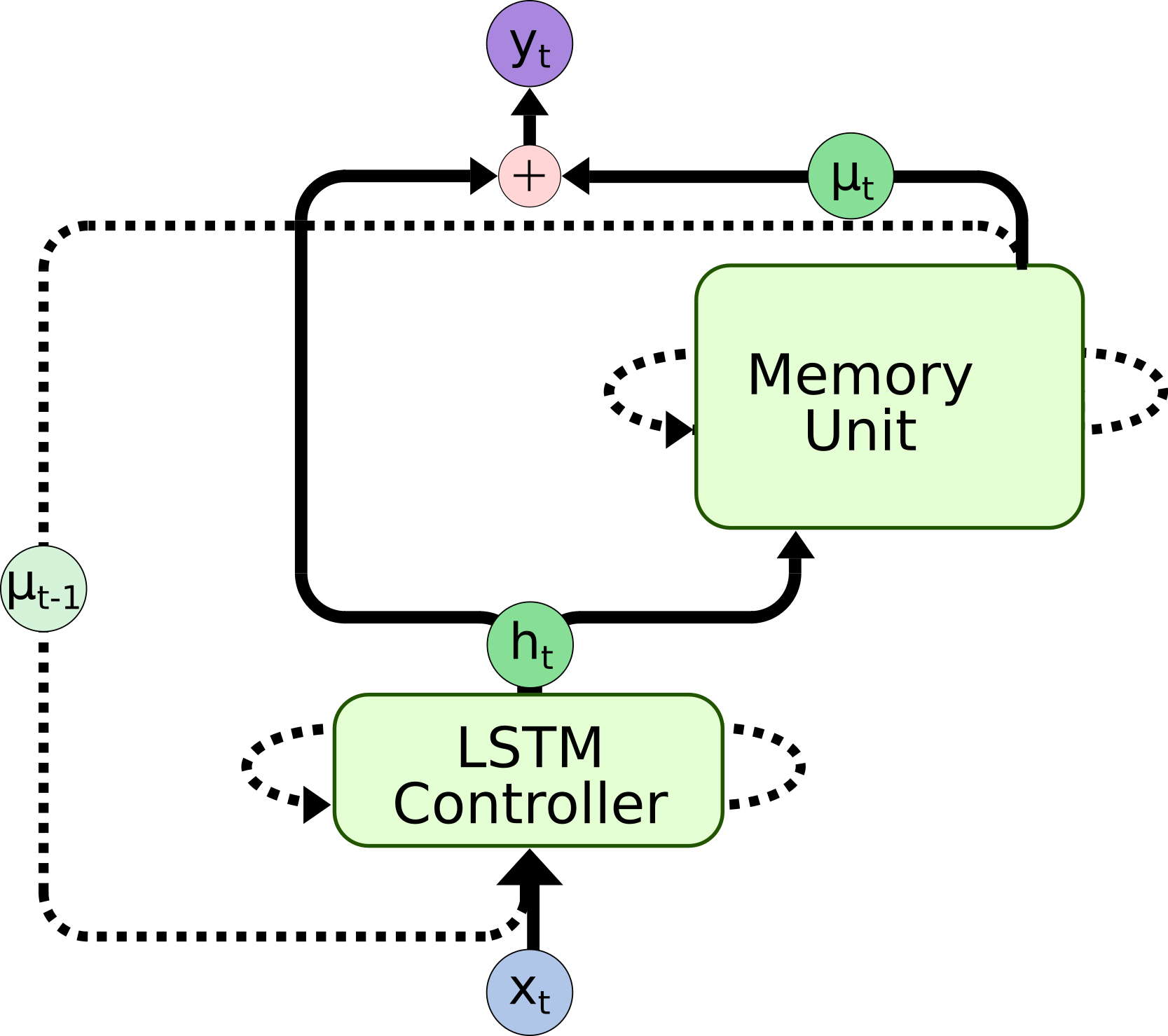}
	\caption[DNC system overview]{System overview of the DNC. The dotted lines illustrate recurrent connections.}
	\label{fig:DNC-system_recurrent}
\end{figure}

\section{Differentiable Neural Computer}

\paragraph{System overview}
The DNC model consists of two main parts, a controller and a memory unit (MU), see Figure \ref{fig:DNC-system_recurrent}. The controller is either a fully-connected ANN or a RNN. It receives at each time step a concatenation of the input signal $\boldsymbol{x}_t \in \mathbb{R}^X$ and the MU output from the last time step $\boldsymbol{\mu}_{t-1}  \in \mathbb{R}^P$. $C$ is the controller output size and $P$ the MU output size. The controller can be considered as a closed function with a set of trainable weight parameters $\theta_{c}$:
\begin{equation}
\boldsymbol{h}_t = \text{Controller}([\boldsymbol{x}_t, \boldsymbol{\mu}_{t-1}], \theta_{c})  \ \raisebox{2pt}{.}
\end{equation}

The purpose of the controller is to manage the MU and additionally to help building the output signal via a weighted bypass connection. The MU receives the controller output $\boldsymbol{h}_t$ as input and contains a set of trainable weight parameters $\theta_{\mu}$ as well:
\begin{equation}
\boldsymbol{\mu}_t = \text{MemoryUnit}(\boldsymbol{h}_t, \theta_{\mu})  \ \raisebox{2pt}{.}
\end{equation}

The output signal of the whole DNC  $\boldsymbol{y}_t \in \mathbb{R}^Y$ is a sum of the weighted controller output and the weighted MU output. 
\begin{equation}
\label{eqn:dnc_output_1}
\boldsymbol{y}_t = \mathbf{W}_h \boldsymbol{h}_t + \mathbf{W}_\mu \boldsymbol{\mu}_t + \boldsymbol{b}_t \ \raisebox{2pt}{.}
\end{equation}

Let $Y$ be the size of the target vector $\boldsymbol{z}_t \in \mathbb{R}^Y$. $\mathbf{W}_h \in \mathbb{R}^{Y\times C}$,  $\mathbf{W}_\mu \in \mathbb{R}^{Y\times P}$ and $\boldsymbol{b}_t \in \mathbb{R}^{Y}$ are trainable weights.

The hyperparameters of the controller network are the number of layers $l$ and the nodes per layer $C_l$.  The hyperparameters of the MU are the width $W$ of the internal memory matrix and the number of locations $N$. There can be multiple read heads $R$ which extract information from the memory matrix. The MU output size is then $P=R\times W$.

\paragraph{Memory Unit}
The MU contains a memory matrix $M_t \in \mathbb{R}^{N\times W}$ which stores information in form of row-wise vectors. For controlling the memory matrix, the input signal from the controller is weighted and divided into different control signals including gates, vectors or keys. They allow to write and read from the memory matrix. 

For writing to or erasing the memory matrix two mechanisms are available to find the location where to manipulate $M_t$: By the least used memory location or content-based addressing. The least used memory location is used to find free space for new information. A \textit{free gate} determines the freeing of used memory to allow forgetting content. Content-based addressing finds the location in $M_t$ which has the lowest cosine-similarity to a given key $k_t \in \mathbb{R}^{W}$. An \textit{allocation gate} determines which mechanism is used and a \textit{write gate} determines the intensity of writing or erasing new information to $M_t$. 

Two mechanisms are also available for reading from $M_t$: A content-based addressing similar to the writing and a temporal linkage mechanism. The temporal linkage mechanism contains a linkage matrix ${LM}_t \in \mathbb{R}^{N \times N}$ which stores the transition of the current write location and the previous location to repeat the sequences in forward direction or with a transposed linkage matrix in backward direction. A \textit{read mode} determines which mode is used. Furthermore, a DNC may have multiple ($R$) \textit{read heads} with such reading mechanisms. The MU's output is a concatenation of the read vectors from all \textit{read heads} $\boldsymbol{\mu}_t  \in \mathbb{R}^{W \times R}$. The DNC is described in detail in the appendix of the original paper~\cite{graves2016hybrid}.         

\section{Analysis of the DNC}
\label{ana}

This section describes the analysis of the DNC with regard to question answering.

\paragraph{Training progress}

We trained multiple models with same parameters but different initializations on different bAbI tasks. 
While some models converges, other do not. This depends only on the initialization. If a model does not converge but learns to solve the task, it is overfitting the training data.

We further analyze the influence of the memory unit. 
The output of the DNC is a weighted sum of the controller output and the MU output. The influence is determinable by exclusively using the memory output and comparing it to the performance when using both. 
While training, we found a strong correlation between a high usage of the MU and a good performance of the model.
When a model does not converges, the memory unit has nearly no influence on finding the correct answer. 

We assume that the direct training signal over the bypass connection between controller and output leads to a fast success in learning to use the controller exclusively. This could prevent learning to use the MU. Additionally, the MU output in the beginning is noisy which could guide the controller to ignore the MU. This could be influenced by the initialization.

\paragraph{Functionality}
The functionality of the DNC can be analyzed by observing the gates which determine the mechanisms to write or read content. We find the DNC to exclusively use content-based reading when answering a question. The reading from the the temporal linkage mechanism (in forward and backward direction) has only little impact on finding the answer. Furthermore, the usage of the different gates for freeing memory space (\textit{free gate}), determine the write mechanism (\textit{allocate gate}) or write intensity (\textit{write gate}) does not seem very meaningful. Figure \ref{sub:func_dnc} in Appendix \ref{app_func} shows a plot of the DNC gates during a bAbI task 1 sample. 

\paragraph{Computing resources}
In comparison to a LSTM, the DNC requires over two times more memory during training. It mainly depends on the memory size of the DNC and the sequence length. A closer analysis exhibits that the linkage matrix and the temporal memory linkage mechanism are the main drivers behind memory consumption. This biggest influence is the linkage matrix of size $N \times N$. The temporal memory linkage mechanism of the DNC with the configuration of the original bAbI experiment accounts for $50\%$ of the total memory consumption. 
In our implementation, the training time per sample is four times higher in comparison to a TensorFlow LSTM with the aforementioned configuration.

\section{Robust and Scalable DNC}

This section describes in detail the two enhancements of the robust and scalable DNC (rsDNC): An efficient memory unit and a more robust training. Additionally, it introduces a bidirectional DNC architecture.

\subsection{Efficient Memory Unit for QA}
For an efficient usage of the model and the scalability to large-scale tasks, less memory consumption is relevant. This allows dealing with larger sequences and bigger batch sizes for faster iterations needed e.g. for feasible hyperparameter tuning. The memory consumption of the DNC is very high compared to other recurrent models. 
As the analysis in Section \ref{ana} shows, the main cause for memory consumption is the temporal memory linkage mechanism.  But the analysis also shows that the DNC does not use them in the bAbI task. This makes sense since restoring sequences is barely necessary for finding the correct answer in a QA task. 

To allow a more efficient usage in QA we used a slim, only content-based memory unit (CBMU). The CBMU has the same features as the DNC's MU but without the temporal memory linkage mechanism. Consequently, the linkage matrix and all related components are removed. The read weightings are only based on the content-based addressing. The write head, memory update and the actual memory reading stay the same. 

The drop in memory consumption depends on the hyperparameters and the sequence length but in this paper, it is between $30\%$ and $70\%$. The computation time is also reduced by $10\%$ to $50\%$.

\subsection{Robust DNC Training}
A more robust training improves the large variance in the training progress within different initializations as well as the slow and unstable convergence behaviour. This makes the training repeatable and reduces the training time. To achieve this, we apply normalization to the DNC and present Bypass Dropout to force a faster memory usage.

\paragraph{DNC Normalization}

Analysis of the DNC shows a high variance in performance between different runs. We approach this issue with a normalization technique to enforce a robust and smooth convergence behaviour. In recent years especially layer normalization (LN) shows performance improvements in ANNs on several applications \cite{ba2016layer, klambauer2017self}. In the DNC setup, it can be applied to the controller as well as the MU. Let $\mu_{t}$ be the mean of a vector $x_t$ and $\sigma_{t}^{2}$ the variance of it. Then the normalization 

\begin{equation} 
\label{eqn:ln}
\text{LN}(\boldsymbol{x}_t) = \boldsymbol{g} \circ \frac{\boldsymbol{x}_{t}-\mu_{t}}{\sqrt{\sigma_{t}^{2}+\epsilon}}+\boldsymbol{b} 
\end{equation}

is applied before each activation function.  The recurrent and current input signals are computed jointly. Each LN has trainable variables, called bias $\boldsymbol{b}$ and gain $\boldsymbol{g}$ for scaling the normalization. 

In the MU we applied LN to the gates, the vectors and the keys separately but this gave no performance increase compared to a joint normalization of all signals. Thus, we apply it after the weighting of the controller output $h_t$ 

\begin{equation}
\label{eqn:advanced_mu_input}
\xi_t = \text{LN}(\boldsymbol{h}_t W_\xi )  
\end{equation} 

and before the vector $\xi_t$ is split into the different control signals. It is applied during training and test times. The drawbacks are increased computation time and memory need due to momentum calculation and the additional gain and bias variables.

\paragraph{Bypass Dropout}

The analysis in Section \ref{ana} shows that convergence behaviour of the DNC depends on the usage of the MU. If the MU strongly impacts the system output, the model achieves a good performance. This insight demands the explicit force of MU usage during training to reach convergence faster and obtain better performance.

To force the MU's influence, the impact of the controller to the output via the bypass connection can be limited. This can be achieved by reducing the connectivity between the controller and the output. 
A reduction of the connectivity by weighting or lower the feature space would be permanent and not adjustable. 
By using dropout in the bypass connection between controller and output is a controllable lowering of the connectivity possible. Dropout is a regularization technique introduced 2014 in \newcite{srivastava2014dropout} to prevent ANNs from overfitting. 
In our use-case, dropout allows a adjustable reduction of the bypass connectivity. It can be controlled with the keep probability and is only used during training. His helps to enforce a faster usage of the MU. We call this technique Bypass Dropout (BD). 

The dropout probability $p$ regularizes the signal flow exactly without permanent declining the controller's functionality. Bypass Dropout is applied to the DNC setup by multiplying a Bernoulli vector $\boldsymbol{r}_t$ to the bypass connection:

\begin{equation}
\def\arraystretch{1.3}
\begin{array}{@{}ll@{}}
\boldsymbol{r}_t \sim \text{Bernoulli}(p)\\
\boldsymbol{y}_t = \mathbf{W}_h (\boldsymbol{h}_t \circ \boldsymbol{r}_t)  + \mathbf{W}_\mu \boldsymbol{\mu}_t  \ \raisebox{2pt}{.}
\end{array}
\end{equation}

\subsection{Bidirectional DNC}

The unidirectional architecture of the DNC makes it hard to handle variable input where for example the question appears in the middle of a text and the full text is relevant. It also prevents a rich information extraction in forward and backward direction in any sequential task. 

Therefore, this work introduces a bidirectional setup to provide complete availability of the input sequence to the model. No more distinction between context and question is necessary.  The model is able to use information from a later point in the input sequence to determine what to store. 
In the bidirectional DNC (BDNC) and rsDNC (BrsDNC) an additional RNN in backward linkage provides a sequential comprehension in both directions.

\begin{figure}[ht]
	\centering
	\includegraphics[scale=0.095]{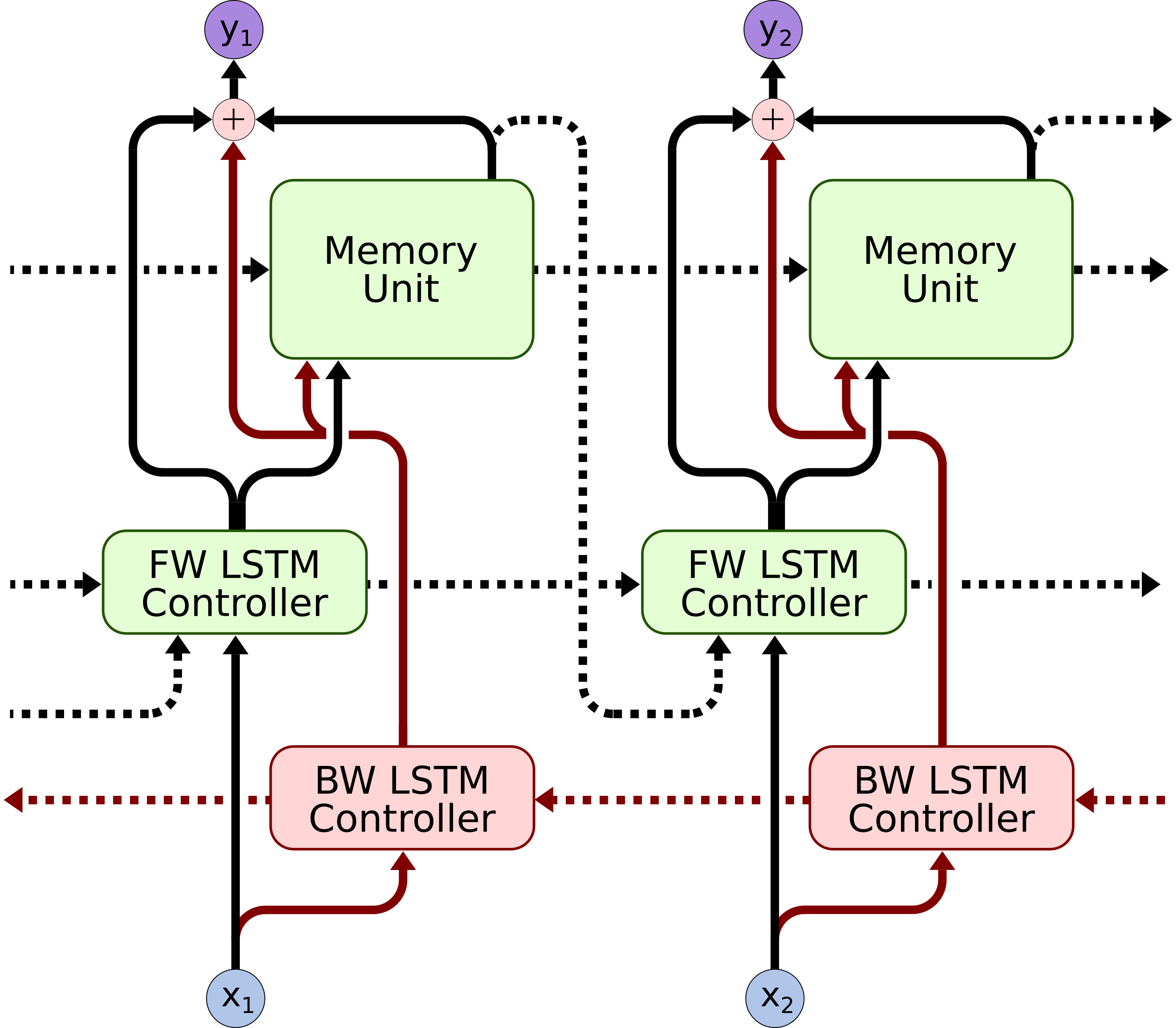}
	\caption[DNC bidirectional controller]{The bidirectional DNC architecture unfolded in time.}
	\label{fig:DNC-bidirectional}
\end{figure}

Due to the recurrent connection from MU to controller, an encapsulated bidirectional controller is not possible, since such a model is not unfoldable in time. The solution presented in this work applies an independent backward-directed recurrent controller which provides an additional input signal to the MU and the output layer, illustrated in Figure \ref{fig:DNC-bidirectional}. Hence, the BrsDNC has two controllers, a forward controller and a backward controller 
\begin{equation}
\def\arraystretch{1.3}
\begin{array}{@{}ll@{}}
\boldsymbol{h}^{fw}_t = \text{ForwardController}([\boldsymbol{x}_t, \boldsymbol{h}^{fw}_{t-1}, \boldsymbol{\mu}_{t-1}], \theta_{c^{fw}})\\
\boldsymbol{h}^{bw}_t = \text{BackwardController}([\boldsymbol{x}_t, \boldsymbol{h}^{bw}_{t+1}], \theta_{c^{bw}}) 
\end{array}
\end{equation}
with independent weights $\theta$ and recurrent connections. The MU receives a concatenation of the two controller outputs
\begin{equation}
\boldsymbol{\mu}_t = \text{MemoryUnit}([\boldsymbol{h}^{fw}_t, \boldsymbol{h}^{bw}_t], \theta_{mu})  \ \raisebox{2pt}{.}
\end{equation}
The output of the BrsDNC system is the sum of the weighted memory output, the weighted backward controller output and  the weighted forward controller output: 
\begin{equation}
\boldsymbol{y}_t = \mathbf{W}_\mu \boldsymbol{\mu}_t + \mathbf{W}_{fwh} \boldsymbol{h}^{fw}_t + \mathbf{W}_{bwh} \boldsymbol{h}^{bw}_t  \ \raisebox{2pt}{.}
\end{equation}
This architecture allows first an independent unfolding of the backward controller and second an unfolding of the forward controller and MU.

\section{Experiments}

The rsDNC and the BrsDNC are evaluated on two datasets, the bAbI task and the CNN RC task.\footnote{All experiments are implemented in TensorFlow \cite{tensorflow2015-whitepaper} and are trained on a single K80 GPU within two to nine days. The source code is available at https://github.com/joergfranke/ADNC.}

\subsection{bAbI Task}

\paragraph{Task description}

The bAbI task is a set of 20 synthetic QA tasks for testing text understanding and logical reasoning \cite{weston2015towards}. Each task has several stories and each story contains a context, one or more questions and the correct answers. There are different sets available but this work only uses the 10k set in English. Each story is pre-processed by removing numbers, transforming all words to lower case and splitting the sequences into word tokens. The whole set contains 156 unique words and three symbols: '?', '!', and '-'. The '-' symbol in an input sequence symbolizes that an answer is requested. From each task 10\% of the samples are used as a validation set. The test set is separate and has 1k questions per task. Inconsistencies in the training set, like a question asked twice, are deleted. The loss metric for the bAbI task is the word error rate (WER), the fraction of incorrectly answered words to all requested words.

\paragraph{Task 16 augmentation}

Many related models struggle with task 16 in the bAbI dataset and only achieve a WER of roughly 50\% \cite{graves2016hybrid,sukhbaatar2015end,xiong2016dynamic}. The task is to find a colour given a name and  name-animal-colour constellations. In the dataset are four different colours, four animals and five names but not equally distributed in the samples. In many samples one or more colours or animals are present multiple times, see training sample of task 16: 

\textsl{Greg is a lion. Julius is a swan. Julius is yellow. Greg is yellow. Brian is a swan. Bernhard is a frog. Brian is white. Lily is a frog. Bernhard is yellow. What colour is Lily? \underline{yellow}}

There are 9k samples in the training set, but in 4181 cases the correct word appears only once in the context. In all other cases, the correct word appears two, three or four times in the context. 
If the model learns only to count the colour words and to answer the colour that appears multiple times, then it is correct in 3449 cases and in 1370 additional cases with a $50/50$ probability. In 4181 cases it is able to guess and has a $1/4$ probability to be correct. This leads to a mean probability over $50\%$ for a correct answer by guessing and counting words. This is a strong local optimum and makes it hard to find a better solution strategy. 

We provide an augmentation of task 16 so that each sample contains different colours and different animals. 
We pretrain a model on augmented task 16 plus all other tasks and then fine-tune it without any augmentation to learn the original distribution of the data. This model is marked as \textit{+aug16}. The augmentation also could be helpful in related models but it is only evaluated with this work.

\paragraph{Training details}

Different enhanced DNC models are evaluated on the bAbI task as well as the rsDNC with Bypass Dropout, DNC Normalization and the CBMU, the bidirectional rsDNC (BrsDNC) and the BrsDNC with augmented task 16 (BrsDNC +aug16). All tasks are trained jointly on all bAbI tasks without any additional information. 

For a direct comparison, the hyperparameters are based on the original paper \cite{graves2016hybrid}. The unidirectional controller has one LSTM layer and 256 hidden units and the bidirectional has 172 hidden units in each direction. Thus both models have about 891k parameters. The MU of all models has 192 locations, a width of 64 and 4 read heads. All models have a dropout probability of 10\%. 

Each word of the bAbI task is encoded as a one-hot vector with the size of 159. An additional sequence mask is used to generate only training signals when an answer is requested, so the outputs of all other time steps are ignored. The output is a  vector of size 159 and normalized with a softmax function. For training, the cross-entropy loss between the prediction vector and the target one-hot vector is minimized. 

Each training uses mini-batches with a batch size of 32. Different-length sequences are padded. The maximum sequence length during training is limited to 800 words. The optimizer is found using a grid search, the underlined options are used: optimization algorithm [SGD, Adam, \underline{RMSprop}] with a learning rate of $[1, \underline{3}, 6]*10^{-[3, 4, \underline{5}]}$ 
and a momentum of 0.9 \cite{kingma2014adam,tieleman2012lecture}. Each model runs five experiments with different initializations. 

\begin{table}[ht!]
\centering
\begin{tabular}{l|r}
\toprule
Model & Mean WER \\ \midrule
DNC*                   & 16.7             $\pm$ 7.6              \\
EntNet*                  & 9.7            $\pm$ 2.6              \\
SDNC*                  & 6.4            $\pm$ 2.5              \\ \midrule
DNC +CBMU              & 17.6            $\pm$ 1.6              \\
DNC +CBMU+Norm         & 13.6            $\pm$ 3.3              \\
DNC +CBMU+BD           & 9.7             $\pm$ 2.9              \\
\textbf{rsDNC} (CBMU+Norm+BD)   & 6.3             $\pm$ 2.7              \\
\textbf{BrsDNC} (bidirectional rsDNC)   & 3.2             $\pm$ 0.5              \\
BrsDNC +aug16                & 0.4             $\pm$ 0.0      \\ \bottomrule             

\end{tabular}
\caption{Mean word error rate of different models and our enhancements to the DNC in the bAbI task. All models are trained jointly on all 20 bAbI tasks at once without information about the actual task. *Result from \cite{graves2016hybrid,henaff2016tracking,rae2016scaling}}
\label{tbl:enhance}
\end{table}

\paragraph{Results}

Table \ref{tbl:enhance} shows the mean word error rate and variance on the test set from all tasks and all models of this work in comparison to the original paper (DNC) \cite{graves2016hybrid} and the EntNet, the best model with reported mean results as far as we know \cite{henaff2016tracking}.

It also shows the performance impact of the different enhancements: DNC Normalization, Bypass Dropout (BD) and the content-based memory unit (CBMU). The use of CBMU leads to a drop in performance but it also reduces the required memory consumption and lowers the standard deviation within different initializations. Both DNC Normalization and BD show a clear performance gain and together they achieve the best performance. 

Experiments on a single bAbI task with step-wise validation provides a closer insight on these results. The usage of the CBMU affects the convergence behaviour only marginally. With use of DNC normalization, the time until convergence is reduced by more than $50\%$. Additional use of BD lowers the mean convergence time further and increases the reliability of convergence. A bidirectional controller speeds up convergence even more and allowing improvements of up to $75\%$ less required iterations. Figure \ref{fig:DNC-conv} in Appendix \ref{app_3} shows the impacts in bAbI task 1.  

In Appendix \ref{app}, Table \ref{fig:babi_eval_mean20} shows the mean word error rate of all tasks and Table \ref{fig:babi_eval_best20} contains the results of the best runs with additional comparison to the RMN \cite{yang2018finding}, the best-reported model in literature as far as we know.

The rsDNC outperforms the DNC from the original paper as well as the jointly trained EntNet. This shows the impact of the normalization and the BD which improves the model performance even without the temporal memory linkage mechanism. This could imply that this mechanism is not important for QA tasks. It also leads to a significant drop in variance which demonstrates that our models are more robust. 

The additional model complexity through the bidirectional design shows clear improvements without requiring more parameters. It outperforms the DNC in terms of mean error as well as variance. The lower variance indicates a very robust model for different random initializations. But through the bidirectional controller, the model has access to information about the question while reading the context similar to the RMN or DMN+. Particularly the performance on task 3, 17 and 19 reaches a new quality in the mean results without any failed tasks. 

The BrsDNC with augmentation of task 16 leads to the best-reported overall results as far as we know. The modifications allow to learn the task correctly and solves it completely. Even when ignoring the task 16 in the results, the performance of this model is better than previous results. This indicates that the corrected task has a positive cross-effect on the learning of the other tasks.  

But the advancements of the rsDNC not only improve the overall performance. The functionality of the MU, as analyzed in Section \ref{ana}, shows more meaningful control signals, see Figure \ref{sub:func_rsdnc} in appendix \ref{app_func}. The \textit{allocate gate} chooses the writing by the least used location when it writes new information and uses content-based writing when it adds supplementary information to existing entries. The \textit{write gate } prevents writing of uninformative words like articles or prepositions. The \textit{free gate } prevents freeing memory when adding new information to not overwrite stored information. The memory influence is maximal when a answer is requested. All of this shows that our advancements probably increase the fundamental functionality of the DNC.  

The training time of our implementation is 40 min per epoch for the rsDNC and 45 min per epoch for the BrsDNC. An epoch of training a DNC takes double the time due to smaller batch sizes and additional computation for the temporal linkage mechanism. 

\subsection{CNN Reading Comprehension Task}

\paragraph{Task description}
The CNN reading comprehension (RC) task is introduced by \newcite{hermann2015teaching} and is based on crawling the online news articles published on the CNN website from April 2007 to April 2015. The dataset contains samples with news article as context and short article summaries as query statements. The articles and query statements are anonymized by replacing all name entities with tokens. This is required since the articles contain name entities, for example celebrities, and the task aims to exploit general relationships between anonymized entities rather than common knowledge. Each article is the source for four queries on average and each query is the finding of a token which is replaced by a placeholder in the query statement.  

The task is divided into training, validation and test sets on a monthly base. The training set contains 90,266 articles and 380,298 samples with a mean length of 775 words. The used pre-processed dataset has all words in lower case and reduced inconsistencies. The vocabulary size of 118,497 is limited to 50k by replacing rarely-seen words with a 'zero' token. The task contains 408 name entity tokens. The loss metric for the CNN is the accuracy, the fraction of all samples with correct words to the number of samples in total.

\paragraph{Training details}
The query and the article are concatenated (the query first) as an input sequence. The sequence is fed to the model word by word and each word is represented as a word vector with a size of 100 and GloVe initialization \cite{pennington2014glove}. The word vector is optimized during training. The target is the correct word represented as the index of a one-hot vector with the size of all name entity tokens. A candidate mask is created which masks out all name entity tokens which are not present in the sample. The last model output predicts the word and is normalized with a softmax function. The cross-entropy loss between the prediction output vector and the target one-hot vector is minimized. 

We use a batch size of 32; different-length sequences are padded starting from the beginning. Two models are evaluated on this task, rsDNC and BrsDNC. All hyperparameters are chosen inspired by related work. The controller is a LSTM with one hidden layer and a layer size of 512 in the unidirectional setup and 384 each in the  bidirectional setup. Both models have a memory matrix with 256 locations, a width of 128 and four read heads. Bypass Dropout is applied with a dropout rate of 10\%. The maximum sequence length during training is limited to 1400 words. The model is optimized with RMSprop with fixed learning rate of 3e-05 and momentum of 0.9. 

\begin{table}[!ht]
\centering
\begin{tabular}{@{}lll@{}}
\toprule
\multicolumn{1}{l|}{Model}                      & valid & test \\ \midrule
\multicolumn{1}{l|}{Deep LSTM Reader}           & 55.0  & 57.0 \\
\multicolumn{1}{l|}{Attentive Reader}           & 61.6  & 63.0 \\
\multicolumn{1}{l|}{\textbf{rsDNC}}                       & 67.5  & 69.0 \\
\multicolumn{1}{l|}{AS Reader}                  & 68.6  & 69.5 \\
\multicolumn{1}{l|}{\textbf{BrsDNC}}                     & 67.1  & 69.8 \\
\multicolumn{1}{l|}{Stanford AR}                & 72.2  & 72.4 \\
\multicolumn{1}{l|}{AoA Reader}                 & 73.1  & 74.4 \\
\multicolumn{1}{l|}{ReasoNet}                   & 72.9  & 74.7 \\
\multicolumn{1}{l|}{GA Reader}                  & 77.9  & 77.9 \\\bottomrule
\end{tabular}
\caption{The validation and test accuracy (\%) of the rsDNC/BrsDNC and others on CNN dataset. }
\label{tbl:cnn-results}
\end{table}

\paragraph{Results}

The results on the CNN RC task are shown in Table \ref{tbl:cnn-results}. Both models achieve passable results compared to task-specific state-of-the-art models. But when focusing on the fact that our model is trained without any adaption to the task like sentence representations or word-windows, without hyperparameter tuning or any other optimization it is a satisfactory result. The bidirectional model outperforms the unidirectional rsDNC, but only on the test set. The training time of our implementation is 14/16 hours per epoch on the rsDNC/BrsDNC. A DNC/BDNC requires up to 35/41h per epoch. The overall training time is 7 epochs on average. The memory savings of the CBMU result in 4 training days instead of 12.   
This shows the model's adaptability to perform other tasks without any task-specific adjustment and its scalability due to a drastic reduction of training time.  

\FloatBarrier

\section{Conclusion}

This work introduces the robust and scalable DNC (rsDNC), an improved DNC applied in QA tasks. It achieves state-of-the-art results on the bAbI task and passable results on the CNN RC task without any task-specific model adaption. 

We show that the rsDNC is more robust and usable in contrast to the vanilla DNC due to a faster and more stable training behaviour as well as less dependence on the initialization. This is caused by the introduced Bypass Dropout and DNC Normalization. These advancements make the DNC easier applicable to other tasks.
Furthermore, we demonstrate the scalability of the rsDNC to a large-scale QA task by introducing a contend-based memory unit. It lowers memory consumption and training time accompanied by only a minimal loss of performance. 
A novel bidirectional architecture improves the contextual accessibility and allows questions at every position in the input sequence. Additionally, we provide a training augmentation for one of the bAbI tasks to solve all of them. 

In further work, the rsDNC architecture could be extended to a sequence-to-sequence model which enables the usage in other NLP tasks.

\bibliography{acl2018}
\bibliographystyle{acl_natbib}


\appendix

\onecolumn

\section{DNC functionality on the bAbI task 1}
\label{app_func}
\rotatebox{90}{
  \stepcounter{figure}
  \begin{minipage}{.96\textheight}
      \begin{minipage}{.48\textheight}
        \centering
        \includegraphics[width=.9\linewidth]{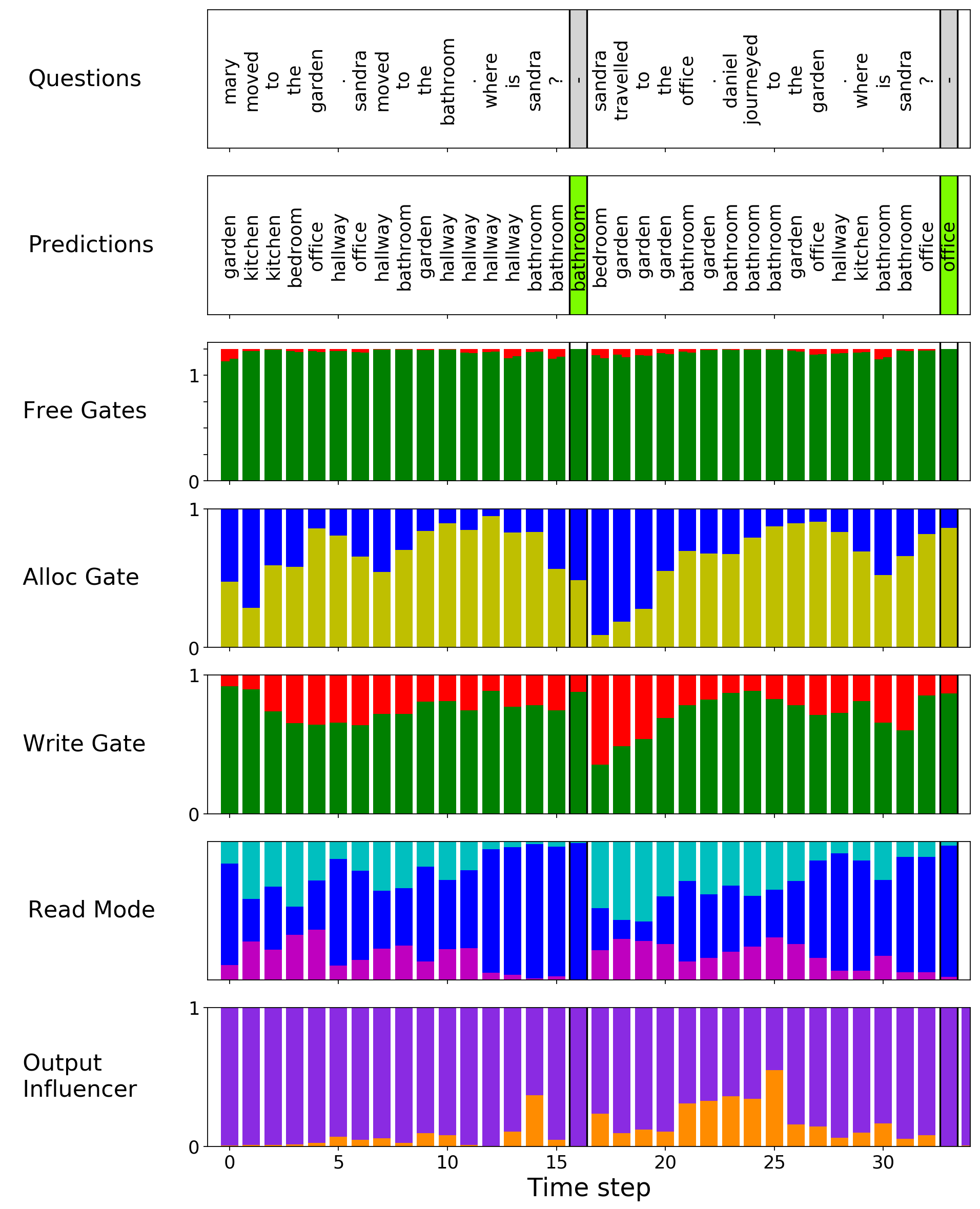}
        \captionof{subfigure}[]{Functionality DNC}
        \label{sub:func_dnc}
      \end{minipage}
      \begin{minipage}{.48\textheight}
        \centering
        \includegraphics[width=.9\linewidth]{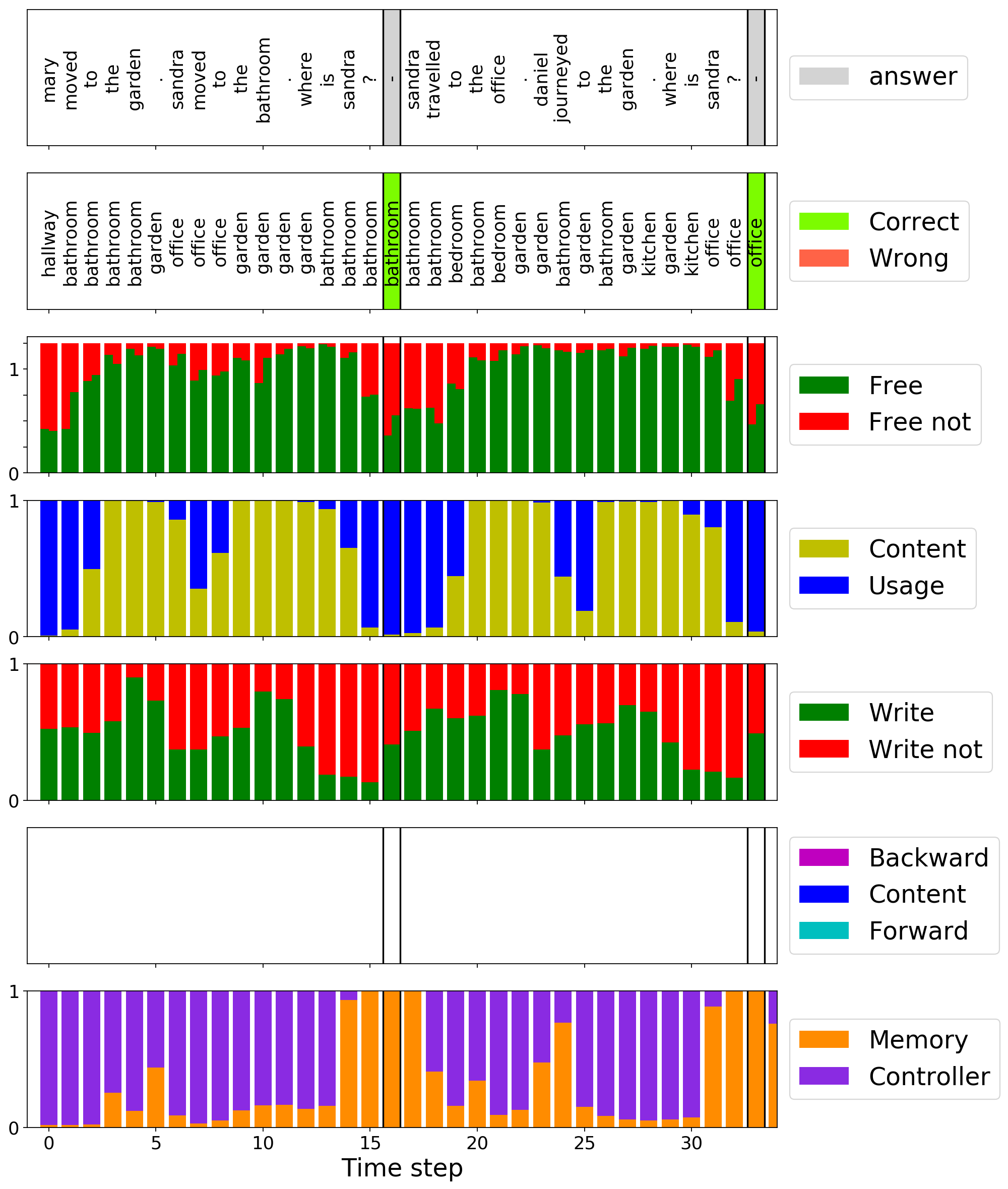}
        \captionof{subfigure}[]{Functionality rsDNC}
        \label{sub:func_rsdnc}
      \end{minipage}
  \addtocounter{figure}{-1}
  \captionof{figure}{The functionality of the DNC's (a) and the rsDNC's (b) gates and the influences in a bAbI task 1 sample. The activity of the gates and read modes are the actual gating values between $0$ and $1$. Influences are normalized between $0\%$ and $100\%$. The gate usage of the rsDNC (b) seems to be more meaningful and the memory influence is higher when answering a question.} 
  \label{fig:func-dnc-rsdnc}
  \end{minipage}
} 

\section{Convergence behaviour on the bAbI task 1}
\label{app_3}

\begin{figure}[ht!]
	\centering
	\includegraphics[width=\textwidth]{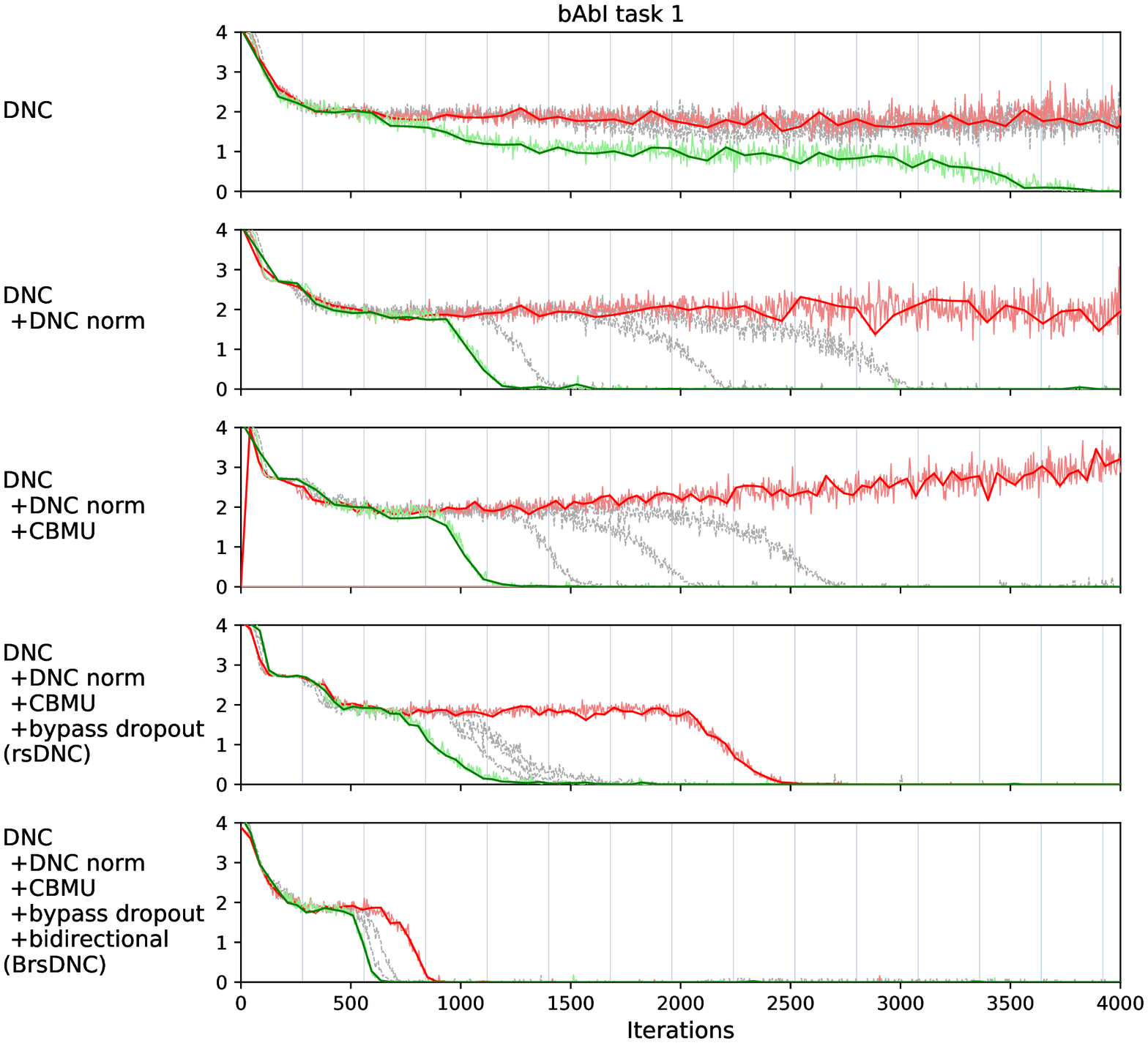} 
	\caption[]{The convergence behaviour of the DNC with different improvements on the bAbI task 1. The plot shows the validation loss for 5 model with the same parameters but different initializations, the red line is the worst model and the green line the best.} 
	\label{fig:DNC-conv}
\end{figure}
\clearpage
\FloatBarrier

\section{Results on the bAbI task}
\label{app}

\begin{table*}[!ht]
\centering
\scalebox{0.82}{
\begin{tabular}{@{}l|llllll@{}}
\toprule
Task                             & DNC             & EntNet                 & SDNC                   & rsDNC                 & BrsDNC                 & \begin{tabular}[c]{@{}l@{}}BrsDNC\\+aug16\end{tabular}          \\ \midrule
1: 1 supporting fact             & 9.0 $\pm$ 12.6  & 0.0 $\pm$ 0.1          & \textbf{0.0 $\pm$ 0.0} & 0.1 $\pm$ 0.0          & 0.1 $\pm$ 0.1          & 0.1 $\pm$ 0.0          \\
2: 2 supporting facts            & 39.2 $\pm$ 20.5 & 15.3 $\pm$ 15.7        & 7.1 $\pm$ 14.6         & 0.8 $\pm$ 0.5          & 0.8 $\pm$ 0.2          & \textbf{0.5 $\pm$ 0.2} \\
3: 3 supporting facts            & 39.6 $\pm$ 16.4 & 29.3 $\pm$ 26.3        & 9.4 $\pm$ 16.7         & 6.5 $\pm$ 4.6          & 2.4 $\pm$ 0.6          & \textbf{1.6 $\pm$ 0.8} \\
4: 2 argument relations          & 0.4 $\pm$ 0.7   & 0.1 $\pm$ 0.1          & 0.1 $\pm$ 0.1          & 0.0 $\pm$ 0.0          & \textbf{0.0 $\pm$ 0.0} & \textbf{0.0 $\pm$ 0.0} \\
5: 3 argument relations          & 1.5 $\pm$ 1.0   & \textbf{0.4 $\pm$ 0.3} & 0.9 $\pm$ 0.3          & 1.0 $\pm$ 0.4          & 0.7 $\pm$ 0.1          & 0.8 $\pm$ 0.4          \\
6: yes/no questions              & 6.9 $\pm$ 7.5   & 0.6 $\pm$ 0.8          & 0.1 $\pm$ 0.2          & 0.0 $\pm$ 0.1          & \textbf{0.0 $\pm$ 0.0} & \textbf{0.0 $\pm$ 0.0} \\
7: counting                      & 9.8 $\pm$ 7.0   & 1.8 $\pm$ 1.1          & 1.6 $\pm$ 0.9          & \textbf{1.0 $\pm$ 0.7} & 1.0 $\pm$ 0.5          & 1.0 $\pm$ 0.7          \\
8: lists/sets                    & 5.5 $\pm$ 5.9   & 1.5 $\pm$ 1.2          & 0.5 $\pm$ 0.4          & \textbf{0.2 $\pm$ 0.2} & 0.5 $\pm$ 0.3          & 0.6 $\pm$ 0.3          \\
9: simple negation               & 7.7 $\pm$ 8.3   & 0.0 $\pm$ 0.1          & 0.0 $\pm$ 0.1          & \textbf{0.0 $\pm$ 0.0} & 0.1 $\pm$ 0.2          & \textbf{0.0 $\pm$ 0.0} \\
10: indefinite knowledge         & 9.6 $\pm$ 11.4  & 0.1 $\pm$ 0.2          & 0.3 $\pm$ 0.2          & 0.1 $\pm$ 0.2          & \textbf{0.0 $\pm$ 0.0} & 0.0 $\pm$ 0.1          \\
11: basic coreference            & 3.3 $\pm$ 5.7   & 0.2 $\pm$ 0.2          & \textbf{0.0 $\pm$ 0.0} & \textbf{0.0 $\pm$ 0.0} & \textbf{0.0 $\pm$ 0.0} & \textbf{0.0 $\pm$ 0.0} \\
12: conjunction                  & 5 $\pm$ 6.3     & \textbf{0.0 $\pm$ 0.0} & 0.2 $\pm$ 0.3          & \textbf{0.0 $\pm$ 0.0} & 0.0 $\pm$ 0.1          & \textbf{0.0 $\pm$ 0.0} \\
13: compound coreference         & 3.1 $\pm$ 3.6   & 0.0 $\pm$ 0.1          & 0.1 $\pm$ 0.1          & \textbf{0.0 $\pm$ 0.0} & \textbf{0.0 $\pm$ 0.0} & \textbf{0.0 $\pm$ 0.0} \\
14: time reasoning               & 11 $\pm$ 7.5    & 7.3 $\pm$ 4.5          & 5.6 $\pm$ 2.9          & \textbf{0.2 $\pm$ 0.1} & 0.8 $\pm$ 0.7          & 0.3 $\pm$ 0.1          \\
15: basic deduction              & 27.2 $\pm$ 20.1 & 3.6 $\pm$ 8.1          & 3.6 $\pm$ 10.3         & \textbf{0.1 $\pm$ 0.1} & \textbf{0.1 $\pm$ 0.1} & \textbf{0.1 $\pm$ 0.1} \\
16: basic induction              & 53.6 $\pm$ 1.9  & 53.3 $\pm$ 1.2         & 53.0 $\pm$ 1.3         & 52.1 $\pm$ 0.9         & 52.6 $\pm$ 1.6         & \textbf{0.0 $\pm$ 0.0} \\
17: positional reasoning         & 32.4 $\pm$ 8    & 8.8 $\pm$ 3.8          & 12.4 $\pm$ 5.9         & 18.5 $\pm$ 8.8         & 4.8 $\pm$ 4.8          & \textbf{1.5 $\pm$ 1.8} \\
18: size reasoning               & 4.2 $\pm$ 1.8   & 1.3 $\pm$ 0.9          & 1.6 $\pm$ 1.1          & 1.1 $\pm$ 0.5          & \textbf{0.4 $\pm$ 0.4} & 0.9 $\pm$ 0.5          \\
19: path finding                 & 64.6 $\pm$ 37.4 & 70.4 $\pm$ 6.1         & 30.8 $\pm$ 24.2        & 43.3 $\pm$ 36.7        & \textbf{0.0 $\pm$ 0.0} & 0.1 $\pm$ 0.1          \\
20: agent’s motivation           & 0.0 $\pm$ 0.1   & \textbf{0.0 $\pm$ 0.0} & \textbf{0.0 $\pm$ 0.0} & 0.1 $\pm$ 0.1          & 0.1 $\pm$ 0.1          & 0.1 $\pm$ 0.1          \\  \midrule
Mean WER:                      & 16.7 $\pm$ 7.6  & 9.7 $\pm$ 2.6          & 6.4 $\pm$ 2.5          & 6.3 $\pm$ 2.7          & 3.2 $\pm$ 0.5          & \textbf{0.4 $\pm$ 0.3} \\
Failed Tasks (\textgreater 5\%): & 11.2 $\pm$ 5.4  & 5.0 $\pm$ 1.2          & 4.1 $\pm$ 1.6          & 3.2 $\pm$ 0.8          & 1.4 $\pm$ 0.5          & \textbf{0.0 $\pm$ 0.0} \\ \bottomrule
\end{tabular}}
\caption[bAbI 20 task mean results]{The mean word error rate (WER) of the different models on the 20 bAbI tasks . All models are trained jointly on all 20 bAbI tasks at once without information about the actual task. Best results in bold.}
\label{fig:babi_eval_mean20}
\end{table*}

\begin{table}[H]
\centering
\scalebox{0.82}{
\begin{tabular}{@{}l|lllllllll@{}}
\toprule
Task                             & DNC          & EntNet       & \begin{tabular}[c]{@{}l@{}}EntNet\\$\dagger$\end{tabular} & \begin{tabular}[c]{@{}l@{}}DMN+\\$\dagger$\end{tabular} & SDNC         & RMN          & rsDNC        & BrsDNC       & \begin{tabular}[c]{@{}l@{}}BrsDNC\\+aug16\end{tabular} \\ \midrule
1: 1 supporting fact             & \textbf{0.0} & 0.1          & \textbf{0.0}                                        & \textbf{0.0}                                      & \textbf{0.0} & \textbf{0.0} & 0.1          & 0.1          & 0.1                                                      \\
2: 2 supporting facts            & 0.4          & 2.8          & \textbf{0.1}                                        & 0.3                                               & 0.6          & 0.5          & 0.8          & 0.5          & 0.6                                                      \\
3: 3 supporting facts            & 1.8          & 10.6         & 4.1                                                 & 1.1                                               & \textbf{0.7} & 14.7         & 2.5          & 2.5          & 1.6                                                      \\
4: 2 argument relations          & \textbf{0.0} & \textbf{0.0} & \textbf{0.0}                                        & \textbf{0.0}                                      & \textbf{0.0} & \textbf{0.0} & \textbf{0.0} & \textbf{0.0} & \textbf{0.0}                                             \\
5: 3 argument relations          & 0.8          & 0.4          & \textbf{0.3}                                        & 0.5                                               & \textbf{0.3} & 0.4          & 1.6          & 0.7          & 0.4                                                      \\
6: yes/no questions              & \textbf{0.0} & 0.3          & 0.2                                                 & \textbf{0.0}                                      & \textbf{0.0} & \textbf{0.0} & \textbf{0.0} & \textbf{0.0} & \textbf{0.0}                                             \\
7: counting                      & 0.6          & 0.8          & \textbf{0.0}                                        & 2.4                                               & 0.2          & 0.5          & 1.5          & 0.3          & 0.6                                                      \\
8: lists/sets                    & 0.3          & 0.1          & 0.5                                                 & \textbf{0.0}                                      & 0.2          & 0.3          & 0.1          & 0.4          & 0.6                                                      \\
9: simple negation               & 0.2          & \textbf{0.0} & 0.1                                                 & \textbf{0.0}                                      & \textbf{0.0} & \textbf{0.0} & \textbf{0.0} & \textbf{0.0} & \textbf{0.0}                                             \\
10: indefinite knowledge         & 0.2          & \textbf{0.0} & 0.6                                                 & \textbf{0.0}                                      & 0.2          & \textbf{0.0} & \textbf{0.0} & \textbf{0.0} & \textbf{0.0}                                             \\
11: basic coreference            & \textbf{0.0} & \textbf{0.0} & 0.3                                                 & \textbf{0.0}                                      & \textbf{0.0} & 0.5          & \textbf{0.0} & \textbf{0.0} & \textbf{0.0}                                             \\
12: conjunction                  & \textbf{0.0} & \textbf{0.0} & \textbf{0.0}                                        & \textbf{0.0}                                      & 0.1          & \textbf{0.0} & \textbf{0.0} & \textbf{0.0} & \textbf{0.0}                                             \\
13: compound coreference         & \textbf{0.0} & \textbf{0.0} & 1.3                                                 & \textbf{0.0}                                      & 0.1          & \textbf{0.0} & \textbf{0.0} & \textbf{0.0} & \textbf{0.0}                                             \\
14: time reasoning               & 0.4          & 3.6          & \textbf{0.0}                                        & 0.2                                               & 0.1          & \textbf{0.0} & 0.1          & 0.1          & 0.5                                                      \\
15: basic deduction              & \textbf{0.0} & \textbf{0.0} & \textbf{0.0}                                        & \textbf{0.0}                                      & \textbf{0.0} & \textbf{0.0} & 0.1          & 0.2          & \textbf{0.0}                                             \\
16: basic induction              & 55.1         & 52.1         & \textbf{0.2}                                        & 45.3                                              & 54.1         & 0.9          & 52.0         & 49.9         & \textbf{0.0}                                             \\
17: positional reasoning         & 12.0         & 11.7         & 0.5                                                 & 4.2                                               & 0.3          & 0.3          & 11.1         & 0.8          & \textbf{0.2}                                             \\
18: size reasoning               & 0.8          & 2.1          & 0.3                                                 & 2.1                                               & \textbf{0.1} & 2.3          & 1.6          & 1.0          & 0.9                                                      \\
19: path finding                 & 3.9          & 63.0         & 2.3                                                 & 0.0                                               & 1.2          & 2.9          & 0.8          & \textbf{0.0} & 0.3                                                      \\
20: agent’s motivation           & \textbf{0.0} & \textbf{0.0} & \textbf{0.0}                                        & \textbf{0.0}                                      & \textbf{0.0} & \textbf{0.0} & 0.2          & 0.1          & \textbf{0.0}                                             \\  \midrule
Mean WER:                      & 3.8          & 7.4          & 0.5                                                 & 2.8                                               & 2.9          & 1.2          & 3.6          & 2.8          & \textbf{0.3}                                             \\
Failed Tasks (\textgreater 5\%): & 2            & 4            & 0                                                   & 1                                                 & 1            & 1            & 2            & 1            & \textbf{0}                                               \\ \bottomrule
\end{tabular}}
\caption[bAbI 20 task best results]{The word error rate (WER) of the best runs on the bAbI 20 task. Best results per row in bold. Models tagged with $\dagger$ are trained on each task individually, the other are trained on all tasks jointly.}
\label{fig:babi_eval_best20}
\end{table}

\end{document}